# APPLICATION OF PARTICLE SWARM OPTIMIZATION TO MICROWAVE TAPERED MICROSTRIP LINES

Ezgi Deniz Ülker[1] and Sadık Ülker[2]

[1]Department of Computer Engineering, Girne American University, Girne, Cyprus
[2]Department of Electrical and Electronics Engineering, Girne American University, Girne, Cyprus

*ABSTRACT*

*Application of metaheuristic algorithms has been of continued interest in the field of electrical engineering because of their powerful features. In this work special design is done for a tapered transmission line used for matching an arbitrary real load to a 50   line. The problem at hand is to match this arbitray load to 50   line using three section tapered transmission line with impedances in decreasing order from the load.  So the problem becomes optimizing an equation with three unknowns with various conditions.  The optimized values are obtained using Particle Swarm Optimization. It can easily be shown that PSO is very strong in solving this kind of multiobjective optimization problems.*

## 1. INTRODUCTION

Metaheuristic algorithms have been applied to engineering optimization problems for a long time. Over the past years an increasing interest for these optimization problems in the field of microwave engineering has increased substantially.  Among these, Particle Swarm Optimization (PSO) is one of the well known and powerful algorithm. It is a population based swarm algorithm and proposed by Kennedy and Eberhart [1]. It is inspired by the behavior of swarms such as fish or birds and found to be very efficient in a wide variety of optimization problems [2,3]. The potential candidates are called particles and PSO depends on moving particles towards the optimal solution to the objective function in the search space [4].

The algorithm can update particles and can be optimized using the following control parameters $c_1$, $c_2$ and $w$.  The term $c_1$ is called *cognitive parameter*, $c_2$ is the *social parameter* and     is called *inertia weight*. $V_{min}$ and $V_{max}$ can be used as *constriction parameters* that define the maximum and minimum positions of a particle in the search space.  PSO algorithm is a parameter dependent algorithm.  The optimal combination of these parameters enables the PSO to reach the best solution rapidly [5].

In PSO, possible particles move in the search space to find the global optimum. This movement is based on proper combinations of control parameters and a replacement formula. Particles also change their positions using their own position called $P_{best}$ (particle best) and swarm's best position $G_{best}$ (global best).  When a better solution is discovered by any particle, all particles improve their positions to this better solution in the search space.

The flowchart for PSO is given in Figure 1. The main steps of the PSO algorithm are given as follows:

Step 1. Initialize the particles $X_i$ with initial random positions in search space, and the velocities of particles $V_i$ in a given range randomly and define these as the best known positions ($P_{best}$) of each particle.
Step 2. Define the objective function $f$ that needs to be optimized.
Step 3. For each particle calculate the distance to the optimal solution called fitness, and apply equation (1):

$$\text{If } f(X_i) < f(P_{best}) \quad P_{best} = X_i \tag{1}$$

Step 4. Select $global_{best}$ among the $P_{best}$
Step 5. If $f(global_{best})$ reaches to the optimal solution, terminate the algorithm. Otherwise; update the velocity $V_i$ and the particles $X_i$ according to given equations:

$$V_i = \omega \cdot V_i + c_1 \cdot (P_{best} - X_i) + c_2 \cdot (global_{best} - X_i) \tag{2}$$

$$X_i = X_i + V_i \tag{3}$$

Step 6. Go to *Step 3*, if stopping criteria are not satisfied.

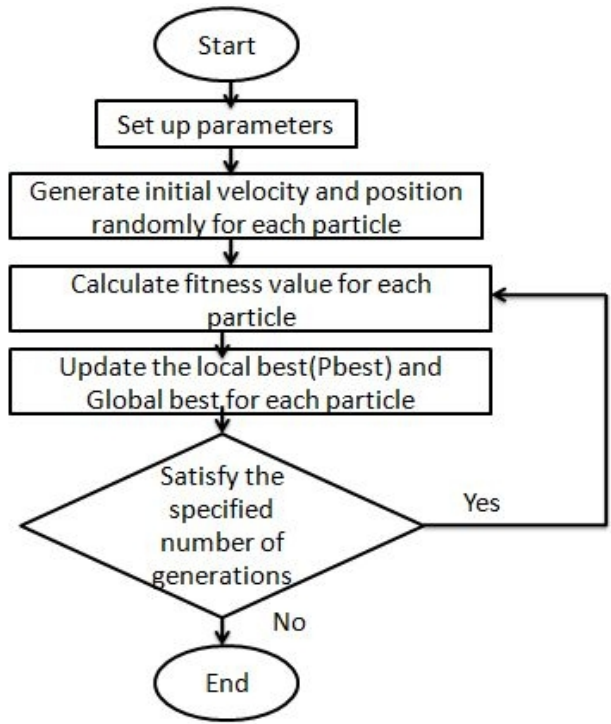

Figure 1: Flowchart of Particle Swarm Optimization [6]

## 2. MICROWAVE PROBLEM

Matching circuits are the essential elements in microwave circuit design. There can be many different methods that can be applied depending on many different considerations. Among the matching circuit types, we can consider shunt stub matching and lumped element matching as popular techniques if the loads are complex loads.

If the matching load is a real load, quarter wave matching networks can also be used. Tapered lines are also used in designing matching network circuits. Among the different ones are Exponential Taper and Klopfenstein Taper [7]. Depending on desired bandwidth in design, proper design can be made with a sacrifice in return loss. A circuit diagram for a microstrip tapered line is shown in Figure 2 below.

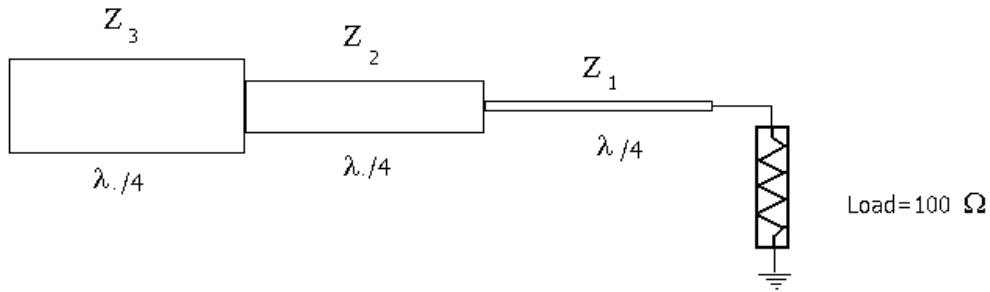

Figure 2. Microstrip Tapered Line

As we can see in the figure, we can cascade couple of sections of high impedance to lower impedance transmission lines of quarter wave long, starting from the load impedance to form a tapered transmission line to achieve matching.

## 3. EXPERIMENTAL WORK

In this case, problem at hand can be thought as of three section tapered transmission line with each transmission line having a different impedance value and each being quarter wavelength long.

The equation that descibes the given problem can be simplified as:

$$f(Z_1, Z_2, Z_3) = \frac{1}{100} \cdot \left[\frac{Z_1 \cdot Z_3}{Z_2}\right]^2 - 50 \qquad (4)$$

We try to make the value of above function 0. This is one constraint, the other constraint that we have is $Z_3 < Z_2 < Z_1$. In order to solve this problem, 100 particles which hold the values of $Z_1$, $Z_2$ and $Z_3$ were created randomly and the PSO algorithm was applied to obtain the optimum values of $Z_1$, $Z_2$ and $Z_3$ that minimizes the value of equation (4).

The impedances were kept between 10 and 120 all the time. Also, as mentioned above, the second criteria was to have the impedances in decreasing order (i.e. the impedance closest to the

Computer Science & Engineering: An International Journal (CSEIJ), Vol. 4, No. 1, February 2014

load was going to have the highest impedance value). The PSO parameter values were as follows: social and cognitive parameters were set to 1.8 and inertia weight was set to 0.7. The particles reached to the optimum values very fast. A sample graph showing this convergence to optimum values (i.e. obtaining values that satisfy (4)) can be shown in Figure 3.

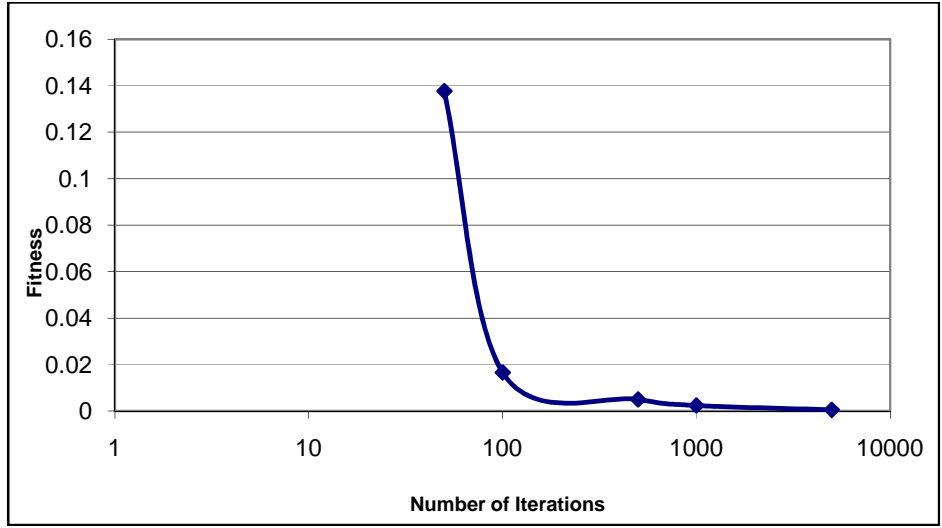

Figure 3. Sample Convergence Graph

As we can see in the graph, in 100 iterations algorithm converges to a good approximate answer and within 10000 iterations the answer can be found. Different runs were done and the sample results are tabulated in Table 1.

Table 1. Obtained Microstrip Impedances from Different Trials

| Trial number | Fitness value | $Z_1$ ( ) | $Z_2$( ) | $Z_3$( ) |
|---|---|---|---|---|
| Trial 1 | 0.000002 | 83.828 | 79.787 | 67.302 |
| Trial 2 | 0.000007 | 86.427 | 55.545 | 45.444 |
| Trial 3 | 0.00051 | 76.604 | 65.646 | 60.595 |
| Trial 4 | 0.000034 | 95.949 | 76.757 | 56.567 |
| Trial 5 | 0.000174 | 78.77 | 67.66 | 60.73 |
| Trial 6 | 0.000022 | 90.012 | 14.131 | 11.101 |
| Trial 7 | 0.000009 | 98.979 | 28.264 | 20.191 |
| Trial 8 | 0.00016 | 96.435 | 30.292 | 22.212 |
| Trial 9 | 0.000005 | 80.797 | 36.922 | 32.313 |

From the table we see that fitness values are all very well in almost all the runs at about 1000 iterations. Moreover the impedances are ordered in decreasing order as we desired. For test purposes the obtained results were simulated using a Microwave Simulator PUFF [8]. For Trial 2 and Trial 3 the responses obtained using the simulator are shown in Figure 4 and Figure 5 respectively.

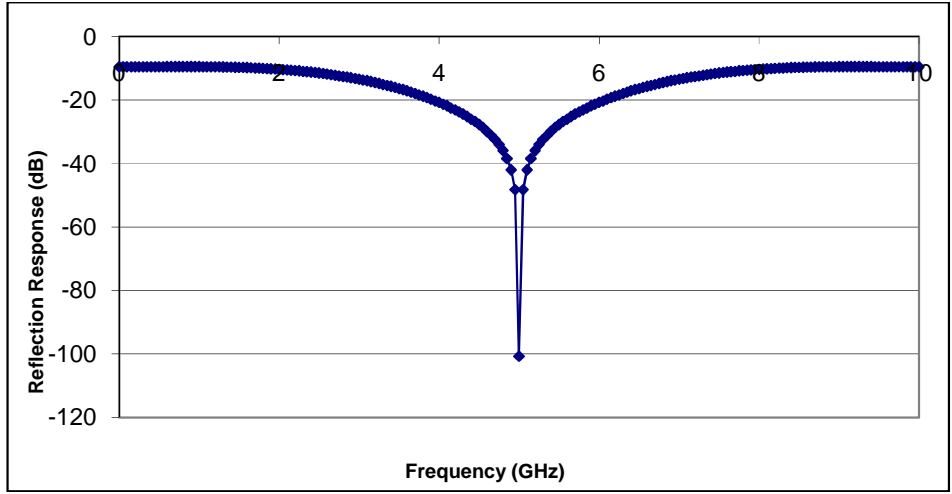

Figure 4. Reflection Response for Trial 2 tapered design (86.427 , 55.545 , 45.444 )

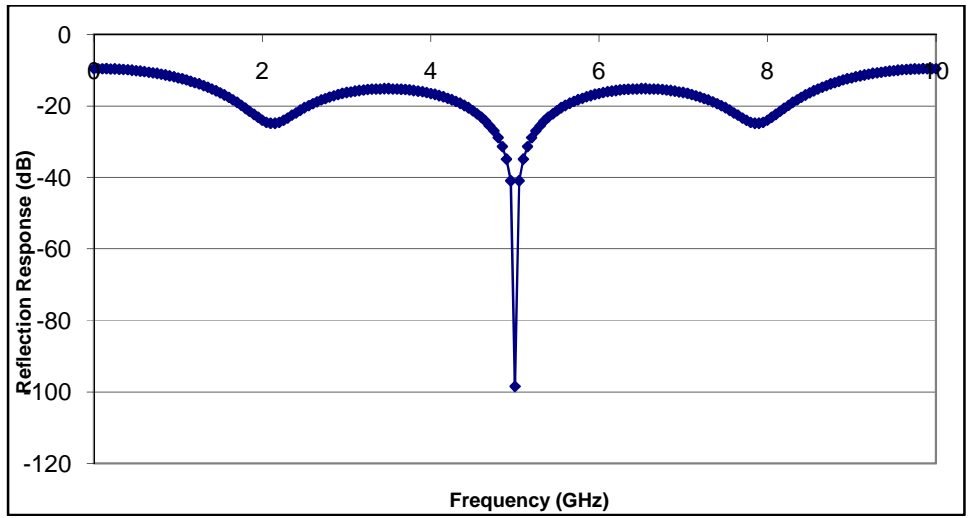

Figure 5. Reflection Response for Trial 3 tapered design (76.604 , 65.646 , 60.595 )

In both cases desired matching is obtained at 5 GHz. Reflection coefficient value being -100 dB indicates almost zero reflection and a good match to a 50 transmission line.

## 4. CONCLUSIONS

In this work, a microstrip tapered transformer design is done using Particle Swarm Optimization. The problem is a multiobjective optimization problem with an objective function that needs to be optimized and with a requirement about the elements of the objective function to be in a decreasing order. The obtained results are verified using PUFF simulator and it can be concluded that PSO finds the solution to this problem in less than 1000 iterations. Simulated results also show that results that were obtained give the matching that is desired. Therefore particle swarm optimization can be very effectively applied for the design of microstrip tapered transformers.

Computer Science & Engineering: An International Journal (CSEIJ), Vol. 4, No. 1, February 2014

**Authors**

**Ezgi Deniz Ulker**
received the B.Sc., M.Sc., and PhD degree in computer engineering from Girne American University, Girne, Cyprus, in 2008, 2010 and 2013 respectively. She worked as a teaching assistant in Girne American University between 2008-2010. She worked as a lecturer in Girne American University from 2010 till 2013. She has been working as Assistant Professor in Girne American University since December 2013. Her research interests include optimization techniques and studying different algorithms and their applications to solve different 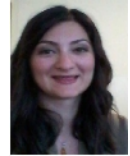
optimization and design problems. She authored and coauthored technical publications in the areas of optimization and metaheuristic algorithms. She is a member of IEEE and IEEE Computer Society.

**Sadik Ulker**
received the B.Sc., M.E. and Ph.D. degree in electrical engineering from University of Virginia, Charlottesville, VA, USA, in 1996, 1999, and 2002 respectively. He worked in the UVA Microwaves and Semiconductor Devices Laboratory between 1996-2001 as a research assistant. Later he worked in UVA Microwaves Lab as a Research Associate for one year. He joined Girne American University, Girne, Cyprus in September 2002. He was a member of electrical and electronics engineering department as an Assistant Professor until 2009. In 2009 he was promoted to Associate Professor status and in 2013 he was promoted to 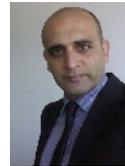
Professor status in the Electrical and Electronics Engineering Department. He has also been the vice-rector of the Girne American University responsible from academic affairs since April 2009. His research interests include Microwave Active Circuits, Microwave Measurement Techniques, Numerical Methods, and Metaheuristic Algorithms and their applications to design problems. He has authored and coauthored technical publications in the areas of submillimeter wavelength measurements and particle swarm optimization technique. He is a member of IEEE and Eta Kappa Nu. He is also the recipient of Louis T. Rader Chairpersons award in 2001.